\documentclass[
  reprint,
  frontmatterverbose,
  nobibnotes,
  amsmath,
  amssymb,
  aps
]{revtex4-2}
\usepackage{array}[=2016-10-06]
\usepackage{graphicx}
\usepackage{bm}
\usepackage{comment}
\usepackage{array}
\usepackage{booktabs}
\usepackage{tabularx}
\usepackage{multirow}
\usepackage[table]{xcolor}
\usepackage{hyperref}
\definecolor{PositionHeader}{HTML}{E8EFF6}
\definecolor{VelocityHeader}{HTML}{EEEAE5}
\definecolor{SummaryShade}{HTML}{F3F4F5}
\newcommand{\bestmean}[2]{\ensuremath{\mathbf{#1}\pm#2}}
\begin{document}

\title{Diffusion models for eye-gaze trajectory generation using position and velocity representations}
\author{Laxman Basnet$^{1}$,
Alexander Szorkovszky$^{2}$,
Pedro G. Lind$^{1,3}$,
Anis Yazidi$^{1,2,4}$,
Shailendra Bhandari$^{1}$}
\email{shailendra.bhandari@oslomet.no}
\affiliation{$^{1}$Department of Computer Science, Oslo Metropolitan University, Oslo, Norway\\
$^{2}$Department of Numerical Analysis and Scientific Computing, Simula Research Laboratory, Oslo, Norway\\
$^{3}$Department of Technology, Kristiania University of Applied Sciences, Oslo, Norway\\
$^{4}$Department of Informatics, University of Oslo, Oslo, Norway}

\begin{abstract}
Eye-tracking data are expensive to collect, requiring specialized hardware and controlled laboratory conditions, and difficult to share due to privacy regulations, which together limit the quantity and diversity of gaze data available for research. We investigate unconditional gaze generation using diffusion models in position and component-wise velocity representations. The models are trained and evaluated on eight-second visual-search trajectories from 28 participants, using a split of 23 training participants and five held-out validation participants, with three training seeds. Evaluation includes kinematic distributions, trajectory geometry, event statistics, and a temporal Wasserstein GAN with gradient penalty as a learned comparator. Across eight position-derived features, mean Jensen–Shannon divergence is $0.016 \pm 0.004$ for the position model and $0.037 \pm 0.002$ for the velocity model, with standard deviations calculated across training seeds. These scores do not establish that representation choice caused the difference. A separate comparison using the same evaluation method shows that the ordering of the models varies across training seeds. The position model also produces smaller feature discrepancies than a kinematic Markov baseline. Differences in saccade counts between real and generated data depend strongly on the velocity threshold used to detect events. The results provide quantitative reference measurements for unconditional gaze generation on this visual-search task and show how evaluation choices affect conclusions about generated gaze.
\end{abstract}

\maketitle

\section{Introduction}
Eye movements offer a powerful window into how humans attend to and process visual information. Where people look, how long they fixate, and how gaze shifts over time carry information about perception, decision-making, and cognitive load, which has made eye-tracking a core method in cognitive science, psychology, and human-computer interaction for decades \cite{Holmqvist2011, rayner1998}. Because gaze reflects both bottom-up saliency and top-down goals, it carries behavioral information that cannot be inferred from static stimuli alone \cite{kummerer2021state}, and it underpins applications from reading research \cite{rayner1998} and usability testing \cite{duchowski2017} to virtual reality \cite{patney2016foveated}, driver monitoring \cite{fridman2018cognitive}, assistive technology \cite{majaranta2014eyetracking}, marketing \cite{wedel2006eyetracking} and clinical diagnostics \cite{leigh2015neurology}.

Meeting this broad demand for gaze data remains difficult. Recordings require specialized hardware, controlled laboratory conditions, and substantial researcher and participant time \cite{Holmqvist2011}, while privacy regulations restrict how recordings can be shared, which limits dataset size, diversity, and reproducibility. Eye movements also vary substantially across individuals even under identical viewing conditions \cite{henderson2014stable}, so the behavioral diversity needed to train robust downstream models is hard to collect at scale. Synthetic data generation is a natural response to this bottleneck. Generative models can supplement scarce recordings, animate virtual or robotic agents with plausible gaze behavior \cite{admoni2017social}, provide a reference distribution for anomaly detection \cite{cogstate2024}, and reduce dependence on costly collection \cite{meijer2024rise}, mirroring synthetic-data efforts in medical imaging \cite{frida2018321} and physiological time series \cite{adib2023ecg}. Doing so is difficult: gaze alternates between low-velocity fixations and rapid saccadic transitions, producing non-stationary, stochastic dynamics that classical Markov and autoregressive models capture only over short horizons \cite{kummerer2021state}. Deep generative alternatives have their own failure modes, with generative adversarial networks (GANs) prone to mode collapse and training instability \cite{pmlr-v70-arjovsky17a, bhandari2025modeling,Lucas2019DontBT}.

Diffusion models provide a non-adversarial alternative. Building on ideas from non-equilibrium statistical physics, Sohl-Dickstein et al. \cite{pmlr-v37-sohl-dickstein15} introduced a forward Markov process that progressively transforms a data distribution into a tractable reference distribution, together with a learned reverse process that reconstructs the data distribution. Ho et al.~\cite{NEURIPS2020_4c5bcfec} developed the DDPM, using a discrete Gaussian forward process and a noise-prediction training objective, while Song et al. \cite{song2021scorebasedgenerativemodelingstochastic} formulated score-based diffusion in continuous time using stochastic differential equations. Because diffusion models do not rely on adversarial minimax optimization, they are generally more stable to train than GANs and have been successfully extended to probabilistic time-series forecasting and imputation \cite{rasul2021autoregressivedenoisingdiffusionmodels, lopezalcaraz2023diffusionbased}. These properties make them well suited to stochastic, temporally structured signals such as eye-tracking data.

Recent diffusion-based approaches to gaze modeling, including DiffGaze \cite{jiao2024diffgaze}, the image-conditioned DiffEye \cite{kara2025diffeye}, and ScanDiff \cite{cartella2025modeling}, have shown that diffusion processes can generate plausible, diverse scanpaths. These models are predominantly stimulus- or task-conditioned, using an image, scene representation, or viewing objective to guide generation. Their evaluation consequently focuses on the agreement between generated gaze behaviour and the corresponding visual stimulus. Less is known about how accurately diffusion models can recover the intrinsic kinematic and temporal structure of continuous gaze signals without explicit stimulus or participant conditioning. We examine these statistics using position- and velocity-space diffusion models.

We investigated this question using two unconditional DDPM formulations trained on the same underlying eye-tracking dataset and visual-search task. Both models use the same participant partition, training and validation window counts, training seeds, and position-derived feature definitions. The position space model generates a two-dimensional gaze-position sequence directly, whereas the velocity-space model generates the two-component gaze-velocity sequences obtained by first-order differencing of the position signal. We first summarize the shared DDPM formulation and the representation-specific preprocessing (Section \ref{sec:diffusion_math}), then describe the denoising architecture and training settings (Section \ref{sec:method}), and evaluate the generated trajectories using kinematic, temporal, and event statistics (Sections \ref{sec:eval_protocol} and \ref{sec:results}). Differences in evaluation configuration limit the extent to which performance differences can be attributed to representation choice. The contribution is an empirical characterization of unconditional gaze generation for this cohort and task.

\section{Background and the dataset}
\label{sec:diffusion_math}
We give a self-contained summary of the DDPM construction. Full derivations of the Gaussian posterior and the variational lower bound are given in Appendices~\ref{app:gaussian_posterior} and~\ref{app:vlb_derivation}.

\paragraph{Forward process and training objective.}
Let $x_0\in\mathbb{R}^D$ be a clean sample, and let $\beta_t\in(0,1)$ denote the variance of the Gaussian noise added at diffusion step $t$. With $\alpha_t=1-\beta_t$ and $\bar\alpha_t=\prod_{s=1}^t\alpha_s$, the DDPM forward process~\citep{NEURIPS2020_4c5bcfec} admits the closed-form marginal
\begin{equation}
x_t=\sqrt{\bar\alpha_t}\,x_0+\sqrt{1-\bar\alpha_t}\,\varepsilon,\quad \varepsilon\sim\mathcal{N}(0,I),
\label{eq:forward_marginal}
\end{equation}
so a noisy sample at any step $t$ can be drawn directly from $x_0$. A network $\varepsilon_\theta(x_t,t)$ is trained to predict the injected noise by minimizing the expected squared $\ell_2$ error:
\begin{equation}
\mathcal{L}_{\mathrm{DDPM}}=\mathbb{E}_{x_0,\varepsilon,t}\bigl[\|\varepsilon-\varepsilon_\theta(x_t,t)\|_2^2\bigr].
\label{eq:simple_loss}
\end{equation}
The injected noise has unit variance at every diffusion step. Comparisons with direct $x_0$-prediction require the corresponding loss weighting for each parameterization. At inference, deterministic DDIM sampling~\citep{DDIM_song2021} ($\eta=0$) integrates the learned reverse dynamics in far fewer steps than full ancestral sampling, without added stochastic noise.

\paragraph{Gaze-sequence formulation.}
For a time-series window $X_0\in\mathbb{R}^{d\times L}$, vectorization to $\mathbb{R}^D$, $D=dL$, recovers Eq.~\eqref{eq:forward_marginal} unchanged. Both models use $d=2$ channels and $L=2{,}000$ samples (8\,s at 250\,Hz), differing only in what the two channels represent: for the position-space model $\mathbf{r}_i=(x_i,y_i)^\top$ denotes the normalized gaze position at sample $i$; for the velocity-space model  $\mathbf{v}_i=(v_{x,i},v_{y,i})^\top$ denotes the corresponding two-component gaze velocity at recording time $\tau_i$. As the diffusion steps $t$ increase, the structured gaze signal is progressively suppressed and replaced by isotropic Gaussian noise. The reverse model must then reconstruct fixation-like intervals and rapid saccadic bursts from noise.

\begin{figure*}[t]
  \centering
  \includegraphics[width=\textwidth]{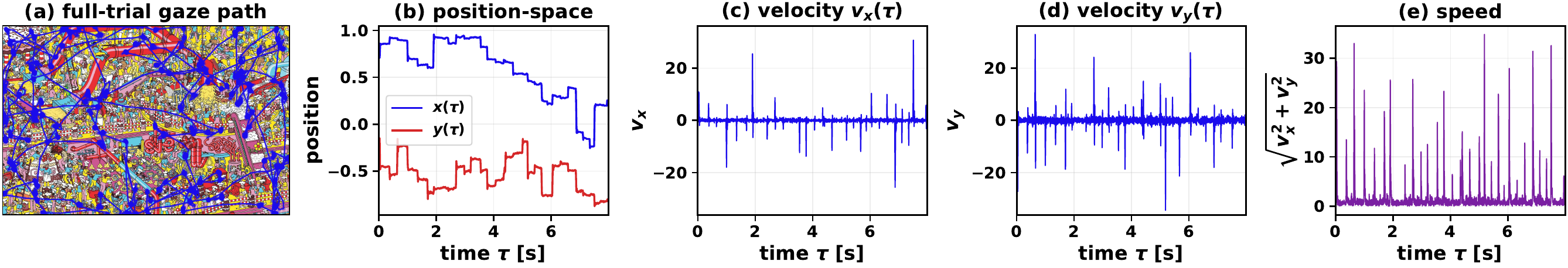}
  \caption{Example recording in both model representations. (a) Gaze path for the full 45\,s trial. (b--e) One 8\,s segment ($L=2{,}000$ samples, $f_s=250$\,Hz) from the same recording: (b) position $x(\tau)$, $y(\tau)$, where fixations appear as plateaus and saccades as steps (input to the position-space model). (c,\,d)~Velocity components $v_x(\tau)$, $v_y(\tau)$ (Eq. \ref{eq:velocity_def}), where saccades become sparse high-amplitude spikes (input to the velocity-space model). (e) Scalar speed $\sqrt{v_x^2+v_y^2}$.}
  \label{fig:dataset_overview}
\end{figure*}

\paragraph{Dataset.} Both models are trained on eye-tracking recordings from a publicly available
dataset \cite{Mathema2026}. The dataset was acquired with an EyeLink Portable Duo system \cite{eyelink_manual} at 1{,}000\,Hz and contains time-stamped binocular gaze coordinates, pupil diameter, and event-state labels (blink, fixation, saccade) for 251 participants across six experimental tasks. Of the available raw channels, both models use only the right-eye horizontal and vertical gaze coordinates as the input signal, since binocular movements are highly correlated under normal viewing and a single eye reduces dimensionality without materially affecting temporal dynamics \cite{Holmqvist2011}.

For both models, analysis is restricted to the Where's Waldo? \citep{handford1987Waldo} visual-search task, in which participants viewed nine visually crowded scenes for 45\,s each while searching for specified characters. The task suits generative modeling: it yields long, continuous, goal-directed trajectories with rich fixation and saccade structure, and the absence of a prescribed scanpath preserves naturalistic inter-individual variability in scanning strategy. Blink-labelled samples are excluded, and recordings are downsampled to $f_s=250$\,Hz by retaining every fourth sample without an explicit anti-alias filter. At this rate, fixations of 150-400\,ms span 37-100 samples and saccades of 20-80\,ms span 5-20 samples. Recordings are segmented into windows of $L=2{,}000$ samples (8\,s), and both models use a participant-level train-validation split so that no participant appears in both partitions (Figure \ref{fig:dataset_overview}).

The position-space model uses 33 participants selected from the 139 available for this task based on data completeness and calibration quality. Pixel coordinates are linearly mapped to $[-1,+1]$ and clipped to $[-1.2,+1.2]$ to accommodate brief off-screen samples. Windows are extracted with an 8\,s window and a 1\,s sliding stride, retaining only windows drawn from contiguous valid runs with less than 30\% of samples exceeding $\pm1.1$ in either axis; participants without any sufficiently long valid run are excluded, reducing the cohort to 28 participants and yielding 4{,}401 valid segments. A fixed participant-level split contains 23 training and 5 validation participants, giving 3{,}265 training and 1{,}136 validation segments for both models.

The velocity-space model uses the same participant partition and windows as the position-space model. Define the sample-to-sample gaze increment as $\mathbf{u}_i =
\mathbf{r}_i-\mathbf{r}_{i-1} =(\Delta x_i,\Delta y_i)^\top,$ where $\Delta x_i=x_i-x_{i-1}$ and $\Delta y_i=y_i-y_{i-1}$. The corresponding horizontal and vertical velocity components are computed by first-order differencing of the normalized coordinates,
\begin{equation}
v_{x,i}=\frac{x_i-x_{i-1}}{\Delta\tau},\qquad
v_{y,i}=\frac{y_i-y_{i-1}}{\Delta\tau},
\label{eq:velocity_def}
\end{equation}
where $i$ denotes the gaze-sample index, $\Delta\tau=1/f_s$, and the first sample is set to zero. Thus, $\mathbf{v}_i=\mathbf{u}_i/\Delta\tau$, with velocity expressed in normalized units/s. Position is reconstructed as $\widehat{\mathbf{r}}_i = \widehat{\mathbf{r}}_{i-1} + \widehat{\mathbf{v}}_i\Delta\tau$, with $\widehat{\mathbf{r}}_0=\mathbf{0}$. Velocities are normalized by $c_v=Q_{0.995}(\{|v_{x,i}|,|v_{y,i}|\})$, the joint training-set 99.5th percentile, and decoded by multiplying by $c_v$. 
The model is evaluated across three independent training seeds (42, 123, and 2024), with 1{,}136 generated windows per seed. Reconstructed trajectories start at \((0,0)\). This choice preserves translation-invariant features, including range, coordinate standard deviation, endpoint displacement, and path length, but prevents evaluation of absolute screen occupancy.
\begin{figure*}[t]
    \centering
    \includegraphics[width=0.95\linewidth]{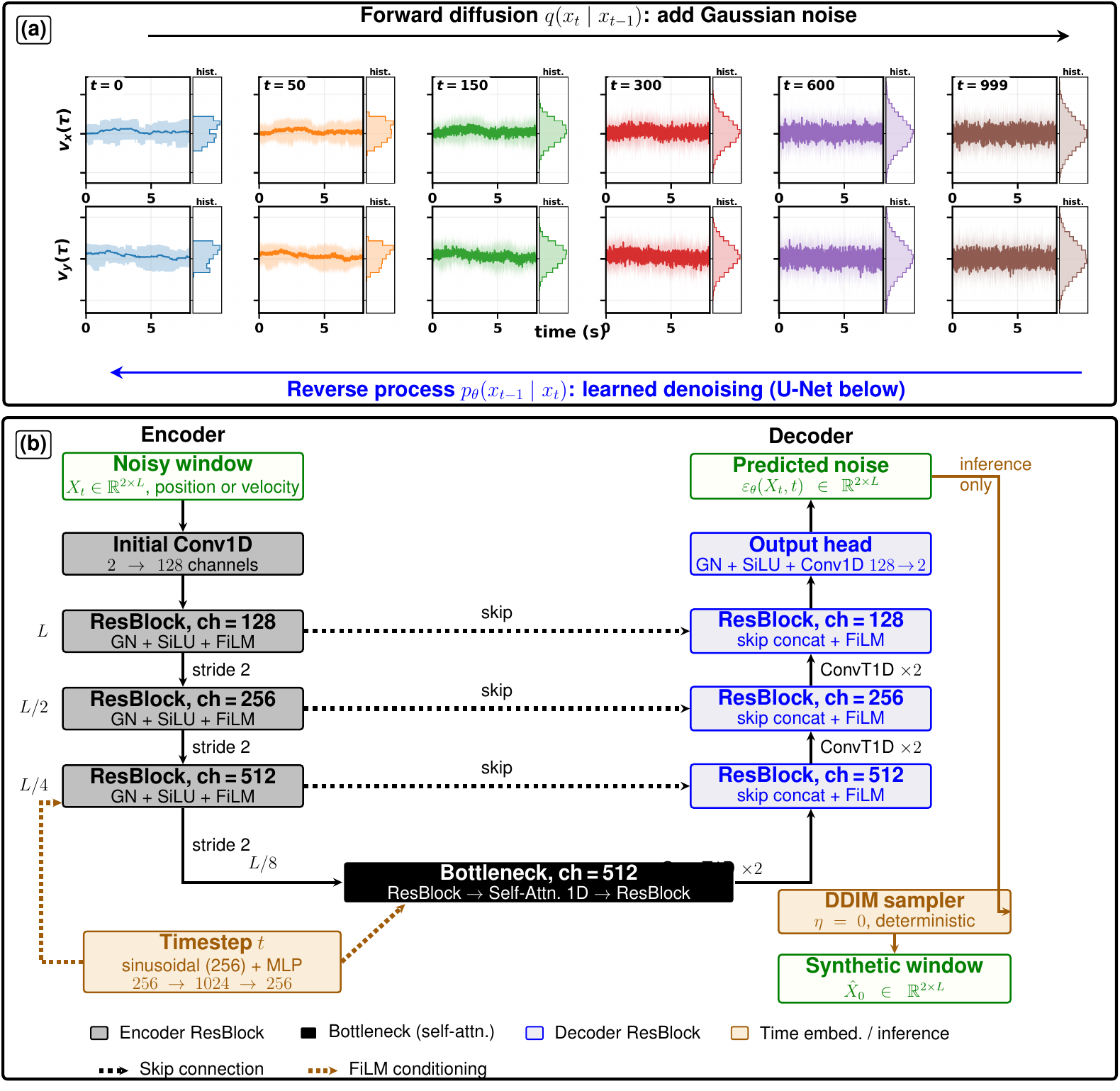}
    \caption{Shared denoising architecture. Position or velocity windows $X_0\in\mathbb{R}^{2\times L}$ are noised to $X_t$, and the 1D U-Net predicts $\varepsilon_\theta(X_t,t)$. Deterministic DDIM sampling ($\eta=0$) produces $\hat X_0$ in the DDPM reference configuration.}
    \label{fig:diffusionmodel}
\end{figure*}

\section{Methods}
\label{sec:method}
The DDPM reference architecture is a one-dimensional U-Net \cite{10.1007/978-3-319-24574-4_28} that receives a noisy two-channel window of shape $(2,L)$ and a diffusion step $t$, and outputs a two-channel noise estimate of the same shape (Figure~\ref{fig:diffusionmodel}). Three encoder stages with channel widths $128\to256\to512$, each comprising a residual block followed by a stride-2 convolution, are followed by a bottleneck with two residual blocks and an interleaved self-attention layer, and a mirrored decoder with transposed-convolution upsampling and skip concatenation from each encoder level, totalling 19{,}350{,}914 parameters. Residual blocks use group normalization and SiLU activations \cite{elfwing2017sigmoidweightedlinearunitsneural} and are conditioned on the diffusion-step embedding via Feature-wise Linear Modulation \cite{perez2018film}, $h'=h\odot(1+\gamma(e_t))+\delta(e_t)$, with the diffusion step encoded using a sinusoidal embedding followed by a multilayer perceptron. The bottleneck self-attention \cite{Vaswani2017} (4 heads, head dimension 32) permits interaction between all bottleneck positions, whose nominal spacing is 32\,ms at 250\,Hz. The DDPM reference uses $T=1{,}000$ diffusion steps under a linear noise schedule from $\beta_1=10^{-4}$ to $\beta_T=2\times10^{-2}$, is optimized with AdamW \cite{loshchilov2018decoupled} under the $\varepsilon$-prediction objective of Eq.~\eqref{eq:simple_loss}, and generates samples via deterministic DDIM sampling \cite{DDIM_song2021} ($\eta=0$). The position-space training settings and DDPM reference configuration are summarized in Appendix Table~\ref{tab:hyperparams}.

\subsection{Position-space training recipe}
The position-space model is trained for 500 epochs with batch size 8, an initial learning rate of $10^{-4}$, weight decay of $10^{-4}$, and gradient clipping at an $\ell_2$ norm of 1.0. The learning rate follows cosine annealing after a 10-epoch linear warm-up \cite{Loshchilov2016}, and training uses automatic mixed precision. Horizontal and vertical flips are applied independently with probability $0.5$, and temporal reversal with probability $0.3$, consistent with the approximate population-level symmetries of the visual-search data. An exponential moving average of the model weights (decay $0.9999$) is used for inference, and trajectories are generated using 100 DDIM steps. The model is trained with three random seeds (42, 123, and 2024) on the fixed 23/5 participant partition. Metrics in Section~\ref{sec:results_position} are reported as mean\,$\pm$\,SD across these runs.

\subsection{Velocity-space training recipe}
The DDPM reference configuration uses 500 epochs with batch size 8, an initial learning rate of $10^{-4}$, weight decay of $10^{-4}$, gradient clipping at an $\ell_2$ norm of 1.0, and a 10-epoch linear warm-up followed by cosine annealing. Training uses the forward relation $\tilde V_t=\sqrt{\bar\alpha_t}\tilde V_0+\sqrt{1-\bar\alpha_t}\,\varepsilon$ derived from Eq.~\eqref{eq:forward_marginal}, together with unweighted $\varepsilon$-prediction MSE, mixed precision, horizontal and vertical flips, and temporal reversal. An exponential moving average of the weights (decay $0.9999$) is used for inference with 100 deterministic DDIM steps. The reference settings are given in Appendix Table~\ref{tab:hyperparams}.

\section{Evaluation protocols}
\label{sec:eval_protocol}
Both models are evaluated on 1{,}136 real and generated validation windows per seed using the same participant partition, training seeds (42, 123, and 2024), and position-derived feature definitions. Their evaluation configurations differ, so the feature table provides a descriptive comparison of the models. A common kernel density estimation (KDE) procedure is used for the separate matched-size path-length and participant-resampling analyses. Unless stated otherwise, reported means and standard deviations summarize three training seeds on one fixed participant partition.

\paragraph{Position-derived features}. We report nine feature distributions for generated positions and positions reconstructed from generated velocities: mean and maximum step speed $s_i=\|\mathbf{u}_i\|_2$ (normalized units/sample), total path length, $x$/$y$-range, $x$/$y$-standard deviation, start-to-end displacement, and fixation ratio. Mean step speed is computed over the $L-1=1{,}999$ increments, so total path length equals $(L-1)$ times mean step speed. Path length is displayed but excluded from the aggregate over eight distinct features. 

The features are compared between $n=1{,}136$ real and $n=1{,}136$ generated validation segments from five held-out participants using the Kolmogorov--Smirnov (KS) statistic \cite{massey1951ks}, JS divergence \cite{lin1991js}, and Wasserstein-1 (W$_1$) distance \cite{villani2009optimal}. W$_1$ carries each feature's own units and is not comparable across features. Values for the position-space model are tabulated separately in Appendix Table~\ref{tab:appendix_position_w1}. KS is interpreted as a distributional discrepancy, since overlapping windows preclude conventional independent-sample significance thresholds. Following the Fr\'echet construction used in FID \cite{Heusel2017}, we compare the mean and covariance of features extracted using a separate 128-dimensional autoencoder trained exclusively on real training trajectories. Its encoder is frozen and shared between both model evaluations under identical preprocessing that centers each trajectory at its first position. Fixations and saccades are identified with the Identification-by-Velocity-Threshold (I-VT) algorithm \cite{salvucci2000identifying} (minimum fixation duration 60\,ms), following established practice for comparative fixation-detection evaluation \citep{andersson2017one, orioma2025adaptive}. We vary the step-speed threshold from $0.010$ to $0.040$ normalized units/sample, using 1{,}136 held-out windows and three seeds per model. The threshold $0.020$ corresponds to 5 normalized units/s at 250\,Hz. Without a validated conversion to degrees/s, the counts are interpreted as threshold-based event labels. We additionally compare the directional and temporal structure of gaze motion using the four-metric framework of \cite{LENCASTRE2023133831}. The four descriptors are step speed $|\mathbf{u}|$, its lag autocorrelation, movement direction $\theta_i=\operatorname{atan2}(\Delta y_i,\Delta x_i)$, and signed turning angle  $\phi_i= \operatorname{wrap}_{[-\pi,\pi]} (\theta_i-\theta_{i-1})$.

The fidelity of the diffusion model is compared against four statistical baselines. The first two of these are first-order time-homogeneous Markov approximations to the velocity and position dynamics, respectively. Following Lencastre et al.~\cite{LENCASTRE2023133831}, 
we approximate the dynamics using a discretized transition matrix $M_{ab}=Pr(c_{i+1}=b\mid c_i=a)$. For the kinematic Markov model, $c_i$ is the cell containing $\mathbf{u}_i$ on a $51\times51$ grid over $[-0.1,0.1]^2$, whereas for the positional Markov model, $c_i$ is the cell containing $\mathbf{r}_i$ on a $32\times32$ grid over $[-1.2,1.2]^2$. The transition matrices are estimated from consecutive training-set transitions and Gaussian-smoothed to reduce sparsity. Synthetic increments are sampled recursively and integrated as
$\widehat{\mathbf{r}}_i=\widehat{\mathbf{r}}_{i-1}+\widehat{\mathbf{u}}_i$,
with $\widehat{\mathbf{r}}_0=\mathbf{0}$. The remaining two baselines draw samples from uniform and Gaussian distributions, respectively.

\paragraph{Velocity diagnostics and participant resampling}. Horizontal and vertical velocity, scalar speed, mean speed profiles, speed autocorrelation, and normalized power spectra \cite{1161901,Percival_Walden_1993} are shown for a representative seed in Figures~\ref{fig:velocity_results} and~\ref{fig:appendix_temporal_diagnostics}. These plots describe marginal and temporal structure separately from the position-derived feature aggregates. The path-length analysis uses identical KDE settings in 100 matched-size draws with $n=127$ per distribution, including participant-disjoint real--real comparisons. A separate analysis resamples complete participant clusters from the five validation participants with replacement for 200 draws, holding generated outputs fixed. The common KDE evaluator reports the velocity-minus-position difference in mean JS over eight distinct features at $n=1{,}136$. The 2.5th--97.5th percentile ranges describe sensitivity to validation-participant composition. They do not measure retraining or new-split uncertainty.

\paragraph{Nearest-neighbour checks.} We compare 1{,}136 generated windows per seed with 3{,}265 training windows, using 1{,}136 validation windows from five held-out participants for calibration. Two trajectory descriptions, relative positions and step velocities, are used. Unusually close samples are defined by the fifth percentile of the real-reference nearest-neighbour distances. The coverage-distance ratio is the median training-to-generated nearest-neighbour distance divided by the median training-to-validation distance.

\section{Results}
\label{sec:results}
\begin{table*}[t]
\centering
\caption{Position-and velocity-space models. Position-derived feature discrepancies, mean $\pm$ SD over training seeds 42, 123, and 2024, with $n=1{,}136$ real and generated validation windows per seed on one fixed 23/5 participant partition. The evaluation configurations differ, so the comparison is descriptive. Nine features are displayed, with redundant total path length excluded from the  eight-feature mean. Lower values indicate smaller discrepancies.}
\label{tab:position_features}
\footnotesize
\begin{tabular}{@{}l cc @{\hspace{20pt}} cc@{}}
\toprule
& \multicolumn{2}{c}{\cellcolor{PositionHeader}\textbf{Position}}
& \multicolumn{2}{c}{\cellcolor{VelocityHeader}\textbf{Velocity}} \\
\cmidrule(lr){2-3}\cmidrule(l){4-5}
\textbf{Feature} & KS & JS & KS & JS \\
\midrule
Mean speed     & \bestmean{0.15}{0.04} & \bestmean{0.03}{0.01}   & $0.19\pm0.03$             & $0.029\pm0.006$ \\
Max speed      & $0.15\pm0.03$        & $0.015\pm0.006$           & \bestmean{0.065}{0.004}   & \bestmean{0.014}{0.002} \\
Path length    & \bestmean{0.15}{0.04} & $0.03\pm0.01$             & $0.19\pm0.03$             & $0.029\pm0.006$ \\
$x$-range      & \bestmean{0.16}{0.02} & \bestmean{0.012}{0.004}   & $0.18\pm0.01$             & $0.066\pm0.002$ \\
$y$-range      & \bestmean{0.12}{0.01} & \bestmean{0.007}{0.002}   & $0.165\pm0.005$           & $0.0533\pm0.0007$\\
$x$-std        & $0.18\pm0.02$        & \bestmean{0.020}{0.003}   & \bestmean{0.17}{0.01}     & $0.037\pm0.001$ \\
$y$-std        & $0.10\pm0.02$        & \bestmean{0.005}{0.001}   & \bestmean{0.086}{0.008}   & $0.026\pm0.002$ \\
Displacement   & $0.14\pm0.01$        & \bestmean{0.0100}{0.0019} & \bestmean{0.117}{0.007}   & $0.021\pm0.002$ \\
Fixation ratio & \bestmean{0.09}{0.06} & \bestmean{0.03}{0.02}     & $0.25\pm0.02$             & $0.046\pm0.004$\\
\midrule
\rowcolor{SummaryShade}
\textbf{Mean (8 features)} & \bestmean{0.14}{0.02} & \bestmean{0.016}{0.004} & $0.154\pm0.005$ & $0.037\pm0.002$ \\
\bottomrule
\end{tabular}
\end{table*}

\subsection{Position-space model}
\label{sec:results_position}
\begin{figure}[t]
    \centering
    \includegraphics[width=\linewidth]{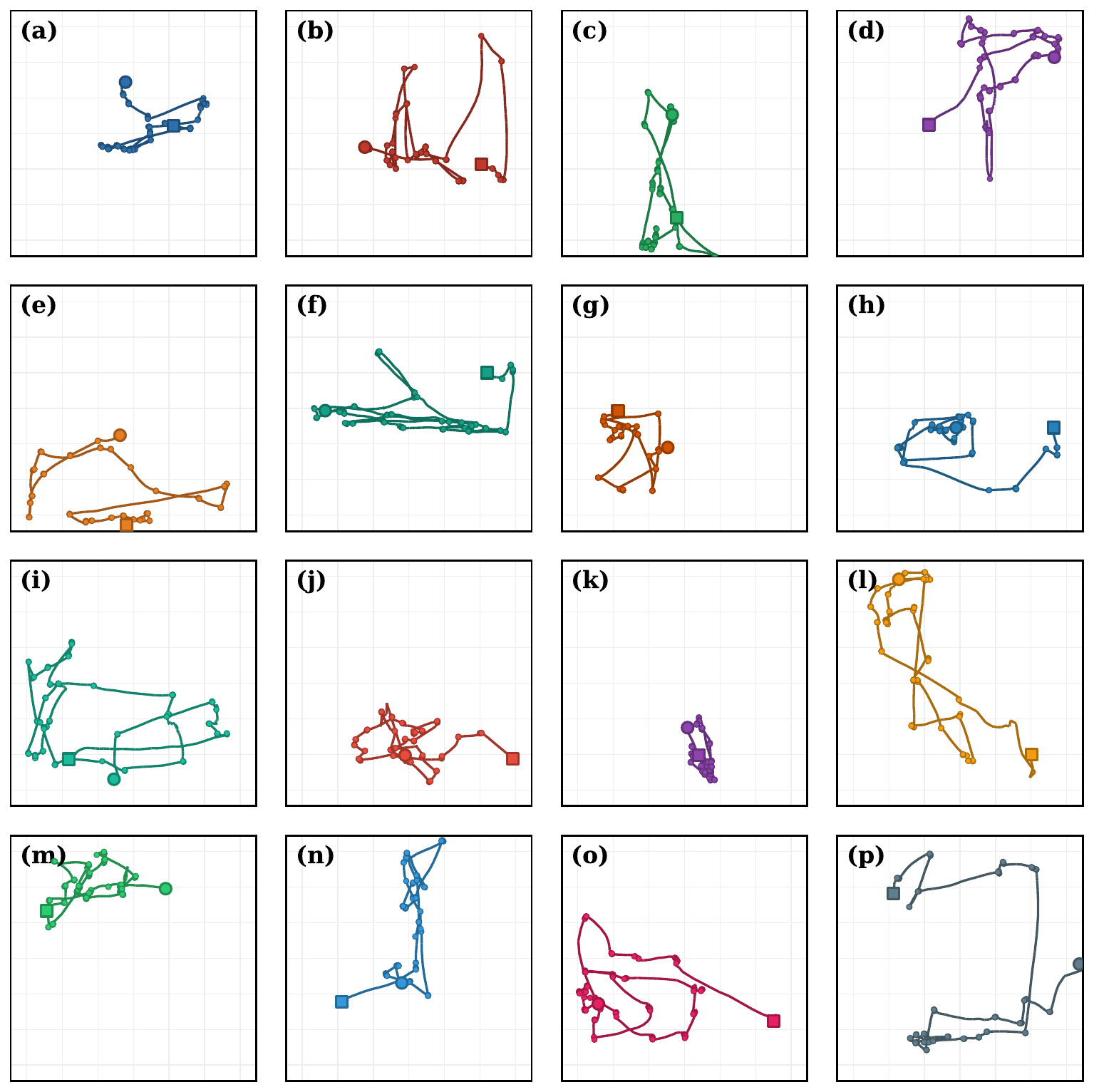}
    \caption{Representative gaze trajectories generated by the position-space DDPM (DDIM, 100 steps, $\eta=0$). Each panel shows the $xy$ path of one generated 8\,s sequence in normalized coordinates.}
    \label{fig:generated_traj}
\end{figure}

Across the three seeds, the best validation loss is $0.0060\pm0.0001$ (reference run: $0.0059$ at epoch 462, training loss $0.0074$ at the same epoch). Trajectories generated by DDIM sampling are visually plausible (Figure~\ref{fig:generated_traj}), spanning a normalized coordinate range consistent with the real validation set and exhibiting clustered, fixation-like dwell periods separated by saccadic jumps.

Table~\ref{tab:position_features} reports the feature-level comparison, while Figure \ref{fig:appendix_feature_distributions} shows the corresponding marginal distributions. Among the nine features, vertical spread ($y$-std, $\mathrm{JS}=0.0051$), vertical range ($0.0067$) and displacement ($0.0100$) are reproduced most closely. Horizontal variability is generally more difficult to match. In particular, $x$-std has the largest KS statistic ($0.1863$ against $0.1009$ for $y$-std) and approximately four times its JS ($0.0206$ against $0.0051$). This $x$-std versus $y$-std ordering holds across all three seeds, whereas the corresponding $x$-range versus $y$-range ordering is not consistent for every seed. The observed anisotropy may reflect greater variability in scanning along the wider horizontal screen axis, although the present experiment does not isolate its cause. Fixation ratio exhibits a different pattern, with the lowest mean KS statistic ($0.0992$) but the largest mean JS divergence ($0.0296$), as well as the largest relative seed variability for both metrics. This combination indicates that the cumulative distributions can agree globally while the corresponding binned densities differ locally, so the low KS value alone should not be interpreted as strong distributional fidelity. Validation loss also did not rank the runs according to sample-level quality: seed 123 had the highest best-validation loss ($0.006147$), but the lowest mean JS over the eight non-redundant features ($0.01225$). With only three seeds, this is an observation rather than evidence of a systematic relationship, but it shows that validation loss alone was not sufficient for assessing generative quality.
\begin{figure}[t]
    \centering
    \includegraphics[width=\linewidth]{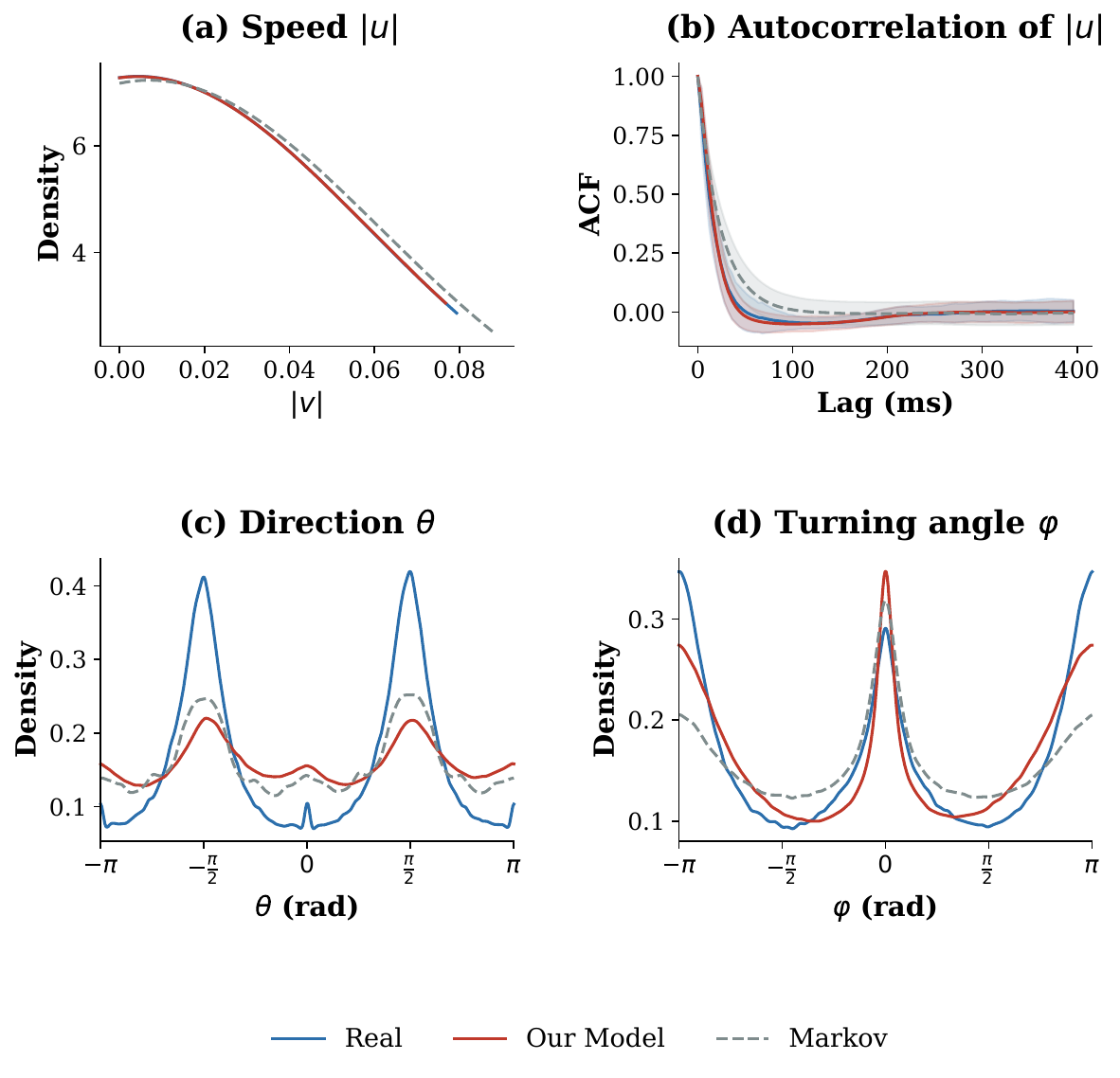}
    \caption {Gaze-trajectory statistics for empirical data, the position-space DDPM, and the kinematic Markov baseline following \citet{LENCASTRE2023133831}: (a) speed $|u|$, (b) autocorrelation of $|u|$, (c) movement direction $\theta$, and (d) turning angle $\phi$.}
    \label{fig:lencastre}
\end{figure}
The sampling ablations in Table~\ref{tab:appendix_ddim_ablation} were computed on a single reference run and should be read as within-run comparisons. The I-VT threshold sweep uses three seeds per model Figure~\ref{fig:ivt_threshold_sensitivity}. At threshold $0.020$, the real data contain a mean of 21.81 {fixations} and 26.79 saccades per window. Across thresholds $0.010$--$0.040$, the position model's relative saccade-count difference ranges from $+71.60\%$ to $-1.43\%$. The velocity model overcounts by approximately $13$--$14\%$ for thresholds $0.015$--$0.040$. Event-count discrepancies therefore depend strongly on the segmentation threshold.

\begin{figure*}[t]
    \centering
\includegraphics[width=0.95\linewidth]{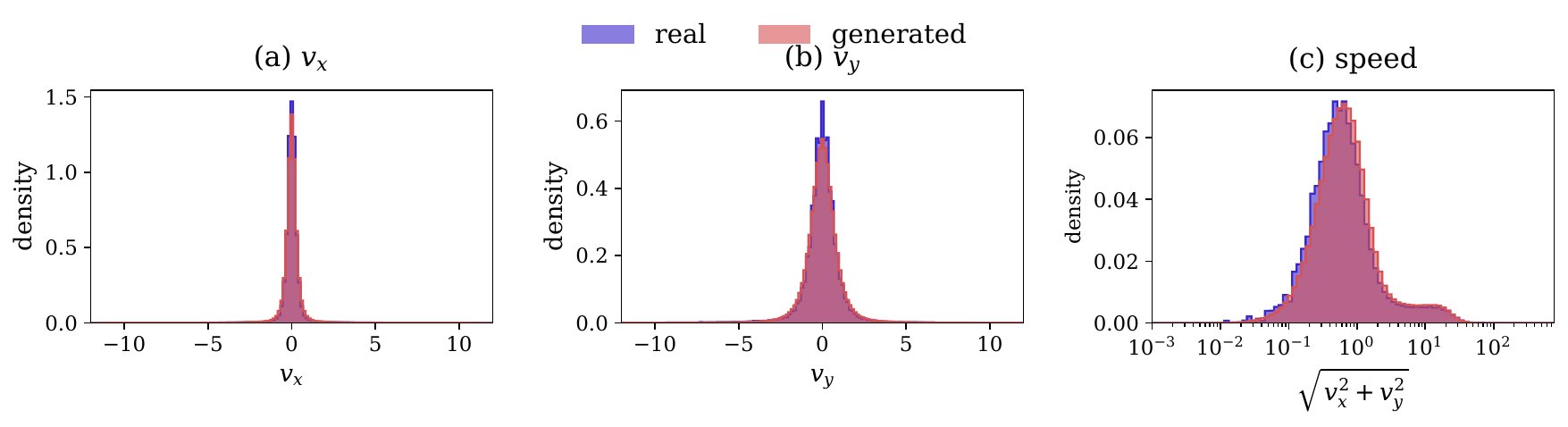}
    \caption{Real and generated gaze-velocity comparison for a representative seed. Pooled empirical distributions over all segments and time points for (a) $v_x$, (b) $v_y$, and (c) scalar speed $\sqrt{v_x^2+v_y^2}$.}
    \label{fig:velocity_results}
\end{figure*}

The position-space model improves on the kinematic Markov baseline for every kinematic feature (mean $\mathrm{KS}=0.138$ against $0.481$; mean $\mathrm{JS}=0.016$ against $0.221$, both over the eight distinct features; Table~\ref{tab:appendix_markov_perfeature}). Figure~\ref{fig:lencastre} compares the position space DDPM and the kinematic Markov baseline with empirical gaze data on the four-metric directional framework. The position-space DDPM closely reproduces the real speed distribution, including the high-speed saccadic tail that the Markov model misses, and both models recover the rapid autocorrelation decay within 50\,ms that reflects the fixation-saccade rhythm. The clearest separation is in turning angle: real gaze has a strong central peak at $\phi=0$, which the position-space model reproduces closely while the kinematic Markov model underestimates it, indicating that the diffusion model better captures this directional dependence than the first-order Markov baseline.

The sampling ablations show a reduction in mean-speed KS from $0.277$ to $0.158$ with EMA weights, and a mean eight-feature KS of $0.135$ for deterministic DDIM ($\eta=0$) versus $0.356$ at $\eta=1$ (Table~\ref{tab:appendix_ddim_ablation}). Increasing the DDIM budget from 100 to 200 steps gives a further, smaller improvement in this reference run.

\begin{table}[t]
\centering
\caption{Independent-encoder Fr\'echet-style distances, mean $\pm$ training-seed SD. Both models use one frozen 128-dimensional real-only encoder and identical first-position-centering preprocessing.}
\label{tab:independent_frechet}
\begin{tabular}{lc}
\toprule
Model & Distance \\
\midrule
Position DDPM & $62.27\pm8.12$ \\
Velocity model & $34.50\pm0.38$ \\
\bottomrule
\end{tabular}
\end{table}

\subsection{Velocity-space model}
\label{sec:results_velocity}
We evaluate the velocity-space model across three independent training seeds (42, 123, and 2024), using the same fixed participant partition and 3{,}265 training and 1{,}136 validation windows as the position-space model. The eight-feature mean JS is $0.0367\pm0.0015$ (Table~\ref{tab:position_features}).
Table~\ref{tab:position_features} reports JS divergences of $0.014\pm0.002$ for maximum step speed, $0.026\pm0.002$ for vertical standard deviation; the signed turning-angle divergence in particular has a small absolute mean but comparatively large relative seed-to-seed variation. Speed profile, autocorrelation deviation, and $0.0215\pm0.0016$ for endpoint displacement. Horizontal and vertical ranges give $0.0661\pm0.0019$ and $0.0533\pm0.0007$, respectively. Path-length JS is $0.0290\pm0.0058$ for the velocity model and $0.0249\pm0.0096$ for the position model. Representative-seed marginal and temporal plots are shown in Figures~\ref{fig:velocity_results} and~\ref{fig:appendix_temporal_diagnostics}.

With identical KDE and matched sample sizes ($n=127$), path-length JS is $0.0670$ for participant-disjoint real--real comparisons, $0.0878$ for the position model, and $0.0497$ for the velocity model. Model values average the three training-seed-specific resampling means over 100 draws. This reference uses five validation participants at $n=127$, so it does not calibrate a full-size feature table at $n=1{,}136$. These descriptive results do not isolate the effect of representation choice.

\subsection{Learned baseline and supplementary diagnostics}
\label{sec:supplementary}
The temporal WGAN-GP \cite{pmlr-v70-arjovsky17a} uses a 17.75\,M-parameter generator with 256-dimensional Gaussian input, four upsampling and convolution blocks, and a strided convolutional critic. Training uses the same 3{,}265 windows, 500 epochs, five critic updates per generator update, $\lambda=10$, and three seeds. Under its separate evaluator, eight-feature, path-length, and turning-angle JS are $0.0765\pm0.0135$, $0.0326\pm0.0275$, and $0.4342\pm0.0099$, respectively (Figure~\ref{fig:gan_velocity_comparison}).

The shared real-only encoder gives Fr\'echet-style distances of $62.27\pm8.12$ and $34.50\pm0.38$ for the position and velocity models, respectively (Table~\ref{tab:independent_frechet}), reversing the descriptive JS ordering. With common KDE settings at $n=1{,}136$, velocity-minus-position mean JS differences are $0.0188$, $0.0257$, and $-0.0018$ for seeds 42, 123, and 2024. Table \ref{tab:fun} reports the 200-draw participant-resampling ranges. Nearest-neighbour checks show no systematic excess below the real-reference fifth-percentile threshold (Table~\ref{tab:nearest_neighbours}), but do not rule out memorization or establish privacy.

\section{Discussion, limitations and conclusions}

The results characterize unconditional gaze generation for a visual-search task. The position-space model  improves on the kinematic Markov baseline across all evaluated feature distributions, extending earlier comparisons of Markov and adversarial models \cite{LENCASTRE2023133831,bhandari2024modeling}. The velocity-space model is evaluated on the same participant partition and position-derived features, while the WGAN-GP provides a learned comparator under a separate evaluation configuration. The matched-size path-length analysis and I-VT threshold sweep show that conclusions about trajectory geometry and event counts depend on the evaluation procedure. These measurements provide empirical reference values for future gaze models \cite{bhandari2025modeling,kara2025diffeye,jiao2024diffgaze,cartella2025modeling}.

Several limitations qualify these findings. Both models are unconditional and cannot condition generation on a specific image, target, task, or participant. The evaluation is restricted to the Where's Waldo? visual-search task and 28 usable participants, with only five held out. Overlapping windows limit the number of independent observations, and validation-based checkpoint selection leaves no untouched test cohort. Participant resampling holds generated outputs fixed and does not measure retraining or new-split uncertainty. Differences in evaluation configuration preclude attributing model differences to representation alone. Real--real calibration is limited to the matched-size path-length analysis, with no full-size reference for the spectral or I-VT diagnostics. The independent encoder has no real--real calibration or assessment of uncertainty from encoder retraining, and it was not used to evaluate the GAN. The learned comparator is adversarial, and attention was not ablated. Decimation without an explicit anti-alias filter may affect velocity and spectral measurements. Sensitivity to blink inclusion has not been quantified. Fixation and saccade counts depend on a velocity threshold without a validated conversion to degrees/s \cite{salvucci2000identifying,andersson2017one,orioma2025adaptive}. Independent downstream utility and privacy remain untested.

The study provides a characterization of unconditional gaze generation in position and velocity representations using kinematic distributions, temporal diagnostics, an independent feature-space distance, participant resampling, and event-threshold sensitivity. The position-space model reduces feature discrepancies relative to the kinematic Markov baseline, while differences between the diffusion models vary with the metric, evaluation procedure, and training seed. Future work should use common training and evaluation configurations, larger independent participant cohorts, stimulus-, task-, and participant-conditioned generation, participant-grouped downstream evaluation with independent targets, and broader assessments of generative diversity and privacy.

\section*{Computational environment}
The diffusion implementation uses Python with PyTorch. We trained the diffusion model on an NVIDIA~A40 GPU. Numerical processing and evaluation used NumPy, SciPy, and Pandas. Figures were produced with Matplotlib.

\section*{Acknowledgments}
We thank OsloMet- Oslo Metropolitan University for institutional support. The authors thank the Research Council of Norway, under the project “Virtual-Eye" (Ref.~335940-FORSKER22).

\bibliography{references}

\appendix

\section{Gaussian posterior of the forward diffusion process}
\label{app:gaussian_posterior}
\begin{figure*}
    \centering
    \includegraphics[width=\linewidth]{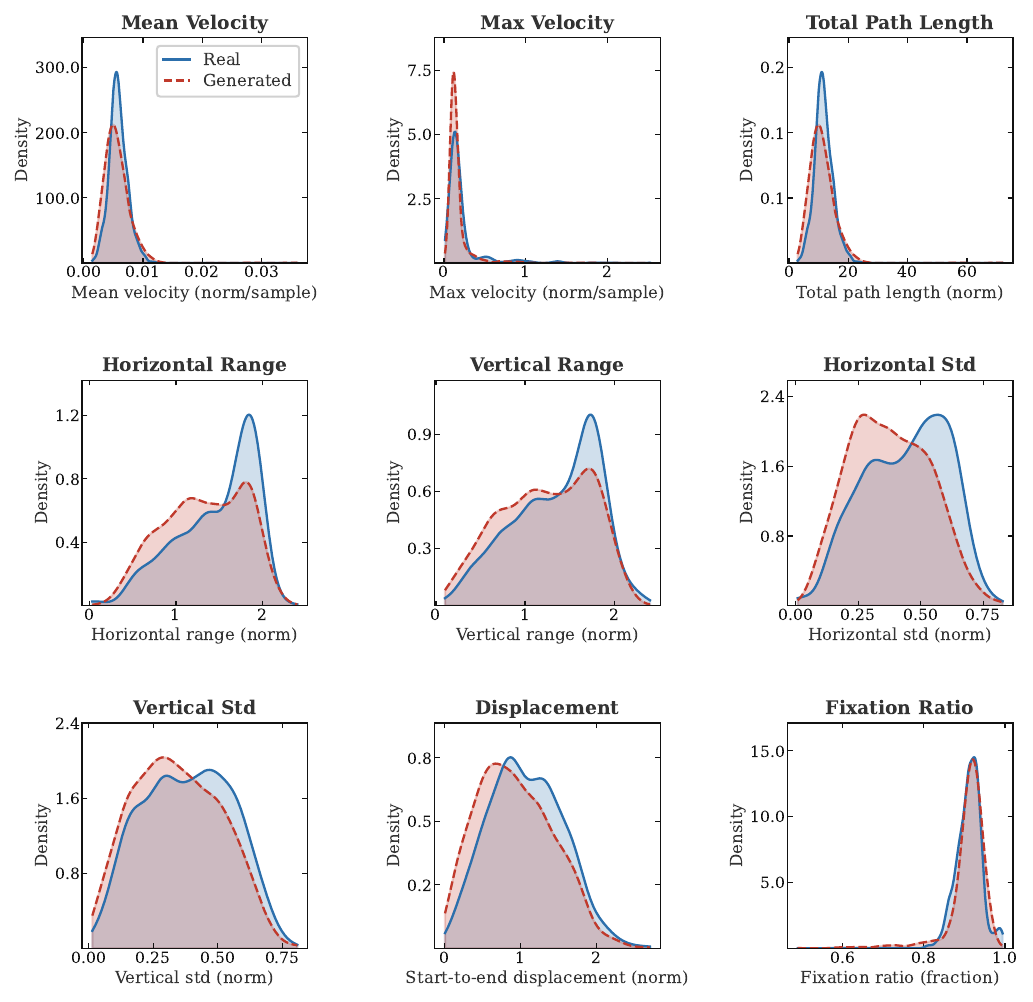}
    \caption{Marginal distributions of the nine kinematic features for real (blue) and generated (red) position-space trajectories ($n=1{,}136$ each); numerical values in Table~\ref{tab:position_features}.}
    \label{fig:appendix_feature_distributions}
\end{figure*}
This appendix gives the closed-form posterior used in Section~\ref{sec:diffusion_math}, following the diffusion-probabilistic formulation of \cite{pmlr-v37-sohl-dickstein15} and the DDPM parameterization of \cite{NEURIPS2020_4c5bcfec}. Let the forward process be $q(x_t\mid x_{t-1})=\mathcal{N}(x_t;\sqrt{\alpha_t}x_{t-1},\beta_tI)$, $\alpha_t=1-\beta_t$, and $\bar\alpha_t=\prod_{s=1}^t\alpha_s$. The marginal at step $t-1$ conditioned on $x_0$ is $q(x_{t-1}\mid x_0)=\mathcal{N}(x_{t-1};\sqrt{\bar\alpha_{t-1}}x_0,(1-\bar\alpha_{t-1})I)$. For $t\ge2$, both factors are non-degenerate Gaussians, so $q(x_{t-1}\mid x_t,x_0)\propto q(x_t\mid x_{t-1})q(x_{t-1}\mid x_0)$ is also Gaussian. At $t=1$, conditioning on $x_0$ gives a degenerate posterior concentrated at $x_0$. Keeping only terms depending on $x_{t-1}$,
\begin{equation}
\begin{aligned}
\log q(x_{t-1}\mid x_t,x_0) &= -\tfrac{1}{2\beta_t}\big\|x_t-\sqrt{\alpha_t}x_{t-1}\big\|^2 \\
&\quad -\tfrac{1}{2(1-\bar\alpha_{t-1})}\big\|x_{t-1}-\sqrt{\bar\alpha_{t-1}}x_0\big\|^2 \\
&\quad +C,
\end{aligned}
\end{equation}
with $C$ independent of $x_{t-1}$. Expanding and collecting powers of $x_{t-1}$ gives a quadratic form $-\tfrac12[A\|x_{t-1}\|^2-2b^\top x_{t-1}]+C'$ with
\begin{equation}
A=\frac{\alpha_t}{\beta_t}+\frac{1}{1-\bar\alpha_{t-1}},
\end{equation}
\begin{equation}
b=\frac{\sqrt{\alpha_t}}{\beta_t}x_t+\frac{\sqrt{\bar\alpha_{t-1}}}{1-\bar\alpha_{t-1}}x_0.
\end{equation}
Completing the square, the covariance is $A^{-1}I=\tilde\beta_tI$ with
\begin{equation}
\tilde\beta_t=\frac{1-\bar\alpha_{t-1}}{1-\bar\alpha_t}\beta_t,
\end{equation}
and the posterior mean is
\begin{equation}
\tilde\mu_t(x_t,x_0)=\frac{\sqrt{\bar\alpha_{t-1}}\beta_t}{1-\bar\alpha_t}x_0+\frac{\sqrt{\alpha_t}(1-\bar\alpha_{t-1})}{1-\bar\alpha_t}x_t,
\end{equation}
so that $q(x_{t-1}\mid x_t,x_0)=\mathcal{N}(x_{t-1};\tilde\mu_t(x_t,x_0),\tilde\beta_tI)$. This is the exact one-step reverse posterior induced by the forward process when the clean sample $x_0$ is known; during sampling, $x_0$ is unknown, and the learned network supplies the required approximation through its noise prediction.
\section{Variational lower bound and the denoising objective}
\label{app:vlb_derivation}
Following the variational formulation of diffusion probabilistic models \cite{pmlr-v37-sohl-dickstein15,NEURIPS2020_4c5bcfec}, the reverse model defines $p_\theta(x_{0:T})=p(x_T)\prod_{t=1}^Tp_\theta(x_{t-1}\mid x_t)$, $p(x_T)=\mathcal{N}(0,I)$. The marginal likelihood $p_\theta(x_0)=\int p_\theta(x_{0:T})\,dx_{1:T}$ is intractable; introducing the forward process as a variational distribution and applying Jensen's inequality gives the evidence lower bound
\begin{equation}
\log p_\theta(x_0)\ge \mathbb{E}_{q}\!\left[\log p_\theta(x_{0:T})-\log q(x_{1:T}\mid x_0)\right].
\end{equation}

\begin{table}
\centering
\caption{Reference hyperparameters of the position/velocity DDPM experiments.}
\label{tab:hyperparams}
\footnotesize
\setlength{\tabcolsep}{4pt}
\renewcommand{\arraystretch}{1.05}
\begin{tabular}{ll}
\hline
\textbf{Setting} & \textbf{Value} \\
\hline
Task & Waldo visual search \\
Participants & 28 \\
Training / validation participants & 23 / 5 \\
Training / validation windows & 3,265 / 1,136 \\
Sampling rate & 250 Hz \\
Window length & 2,000 samples (8 s) \\
Network & 1D U-Net \\
Channel widths & 128, 256, 512 \\
Parameters & 19,350,914 \\
Conditioning & FiLM \\
Self-attention location & Bottleneck \\
Attention heads / head dimension & 4 / 32 \\
Diffusion steps & 1,000 \\
Noise schedule & Linear \\
Noise variance endpoints & $10^{-4}$ to $2\times10^{-2}$ \\
Optimizer & AdamW \\
Learning rate / weight decay & $10^{-4}$ / $10^{-4}$ \\
Batch size / epochs & 8 / 500 \\
Learning-rate schedule & Cosine \\
Warmup & 10 epochs \\
Gradient clipping & 1.0 \\
Mixed precision & Enabled \\
Spatial augmentation & Horizontal/vertical flips \\
Temporal augmentation & Time reversal \\
Training seeds & 42, 123, 2024 \\
EMA decay & 0.9999 \\
Training objective & Noise-prediction MSE \\
Loss weighting & Unweighted \\
Checkpoint selection & Minimum validation MSE \\
Sampler & DDIM \\
Sampling steps / $\eta$ & 100 / 0 \\
Generated windows & 1,136 per seed \\
\hline
\end{tabular}
\end{table}
Substituting the Markov factorizations of $p_\theta$ and $q$, the negative bound decomposes as
\begin{equation}
\begin{aligned}
\mathcal{L}_{\mathrm{vlb}}(x_0)
= D_{\mathrm{KL}}\!\left(q(x_T\mid x_0)\,\|\,p(x_T)\right) \\+ \sum_{t=2}^{T}\mathbb{E}_{q(x_t\mid x_0)}\!\left[
D_{\mathrm{KL}}\!\left(q(x_{t-1}\mid x_t,x_0)\,\|\,p_\theta(x_{t-1}\mid x_t)\right)\right] \\ - \mathbb{E}_{q(x_1\mid x_0)}\!\left[\log p_\theta(x_0\mid x_1)\right].
\end{aligned}
\end{equation}
with $L_T=D_{\mathrm{KL}}(q(x_T\mid x_0)\|p(x_T))$, $L_{t-1}=D_{\mathrm{KL}}(q(x_{t-1}\mid x_t,x_0)\|p_\theta(x_{t-1}\mid x_t))$ for $t=2,\dots,T$, and $L_0=-\log p_\theta(x_0\mid x_1)$. The intermediate terms therefore train the learned reverse kernels to match the exact Gaussian posteriors of Appendix~\ref{app:gaussian_posterior}. Using the noise-prediction parameterization introduced by
\cite{NEURIPS2020_4c5bcfec}, with $x_t=\sqrt{\bar\alpha_t}x_0+\sqrt{1-\bar\alpha_t}\varepsilon$, the network is trained to recover $\varepsilon$ from $(x_t,t)$, giving the simplified objective $\mathcal{L}_{\mathrm{DDPM}}=\mathbb{E}_{x_0,t,\varepsilon}[\|\varepsilon-\varepsilon_\theta(x_t,t)\|_2^2]$ used in Eq.~\eqref{eq:simple_loss}. For the velocity-space model, $x_0$ is replaced by the normalized window $\tilde V_0\in\mathbb{R}^{2\times L}$, and the network predicts $\varepsilon_\theta(\tilde V_t,t)$ accordingly; for the position-space model, $x_0$ is the normalized coordinate window directly.

\begin{figure*}
    \centering
\includegraphics[width=\linewidth]{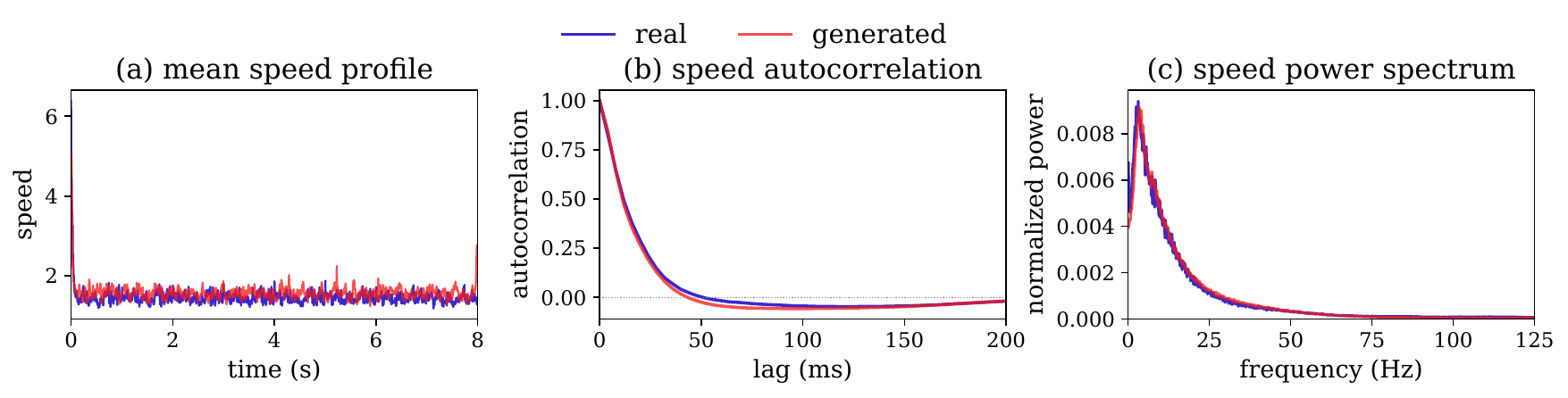}
    \caption{Velocity-space temporal diagnostics for a representative seed. (a) Mean speed profile across segments. (b) Average speed autocorrelation. (c) Normalized speed power spectrum.}
    \label{fig:appendix_temporal_diagnostics}
\end{figure*}

\begin{figure*}[t]
\centering
\includegraphics[width=\textwidth]{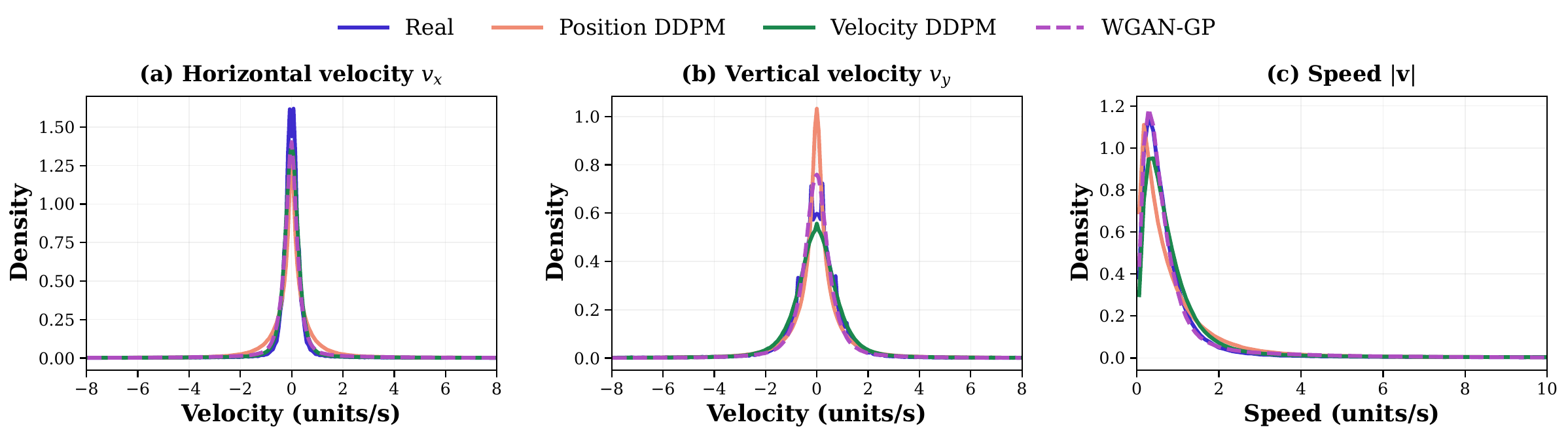}
\caption{Representative-seed velocity-distribution comparison for real trajectories, the position DDPM, velocity model, and WGAN-GP. Horizontal velocity, vertical velocity, and speed are pooled over 1{,}136 generated eight-second trajectories per model and the same real validation set. Position-model velocities are obtained by first-order differencing.}
\label{fig:gan_velocity_comparison}
\end{figure*}
\section{Implementation details and hyperparameters}
\label{app:implementation}
Table~\ref{tab:hyperparams} summarizes the position-space training settings and the DDPM reference configuration, together with the data partition shared by both model evaluations.
\begin{table}
\centering
\caption{Kinematic Markov baseline \cite{LENCASTRE2023133831}: real-data feature means and per-feature agreement with real held-out trajectories. Seed variation is not reported for the Markov baseline. Compared over the eight shared non-redundant features, the position-space DDPM reduces mean KS by $71.4\%$ and mean JS by $93.0\%$. It also improves on the baseline for every individual
feature.}
\label{tab:appendix_markov_perfeature}
\begin{tabular}{lccc}
\toprule
\textbf{Feature} & \textbf{Real }$\mu$ & \textbf{KS} & \textbf{JS} \\
\midrule
Mean speed   & 0.0058 & 0.731 & 0.347 \\
Max speed    & 0.2095 & 0.460 & 0.355 \\
$x$-range       & 1.4699 & 0.589 & 0.225 \\
$y$-range       & 1.3502 & 0.348 & 0.108 \\
$x$-std         & 0.4405 & 0.339 & 0.112 \\
$y$-std         & 0.3815 & 0.157 & 0.049 \\
Displacement    & 1.0730 & 0.435 & 0.155 \\
Fixation ratio  & 0.9121 & 0.790 & 0.416 \\
\midrule
\textbf{Mean} &  & \textbf{0.481} & \textbf{0.221} \\
\bottomrule
\end{tabular}
\end{table}
\section{Additional results}
\label{app:additional_results}
This appendix provides the detailed measurements supporting Section~\ref{sec:results}. Feature distributions and the I-VT threshold sweep are evaluated across three training seeds on one fixed participant partition. Sampling ablations use a single reference run. The additional tables report the matched-size real-data reference, participant resampling, and nearest-neighbourchecks.
\paragraph{Feature-level detail.} Table~\ref{tab:appendix_markov_perfeature} compares the position-space DDPM with the kinematic Markov baseline over the eight features. Averaged over these features, the DDPM reduces the mean KS statistic from $0.481$ to $0.14\pm0.02$ and the mean JS divergence from $0.221$ to $0.016\pm0.004$, corresponding to reductions of $71.4\%$ and $93.0\%$, respectively. The improvement also holds feature by feature, rather than arising from only one or two favorable quantities.

Table~\ref{tab:appendix_position_w1} reports the corresponding Wasserstein-1 distances. Because W$_1$ retains the physical scale of each feature, its values should not be compared directly across rows with different units. In particular, the larger W$_1$ for total path length reflects its scaling by $1{,}999$ relative to mean speed, rather than poorer agreement in distributional shape. Figure~\ref{fig:appendix_feature_distributions} shows the corresponding real and generated marginals and provides a visual counterpart to the KS, JS, and W$_1$ statistics.
\begin{table}
\centering
\caption{Position-space model: Wasserstein-1 distance per kinematic feature, mean\,$\pm$\,SD over three seeds ($n=1{,}136$ per seed). W$_1$ is reported in each feature's own units and is therefore not comparable across rows; total path length is $1{,}999\times$ mean speed by construction, which is why its W$_1$ is correspondingly larger.}
\label{tab:appendix_position_w1}
\begin{tabular}{lc}
\toprule
\textbf{Feature} & \textbf{W$_1$} \\
\midrule
Mean speed     & $0.0006\pm0.0002$ \\
Max speed      & $0.0379\pm0.0075$ \\
Total path length & $1.2195\pm0.3671$ \\
$x$-range         & $0.1355\pm0.0254$ \\
$y$-range         & $0.1071\pm0.0056$ \\
$x$-std           & $0.0631\pm0.0065$ \\
$y$-std           & $0.0361\pm0.0038$ \\
Displacement      & $0.1390\pm0.0204$ \\
Fixation ratio    & $0.0163\pm0.0114$ \\
\bottomrule
\end{tabular}
\end{table}
\begin{figure*}
\centering
\includegraphics[width=\linewidth]{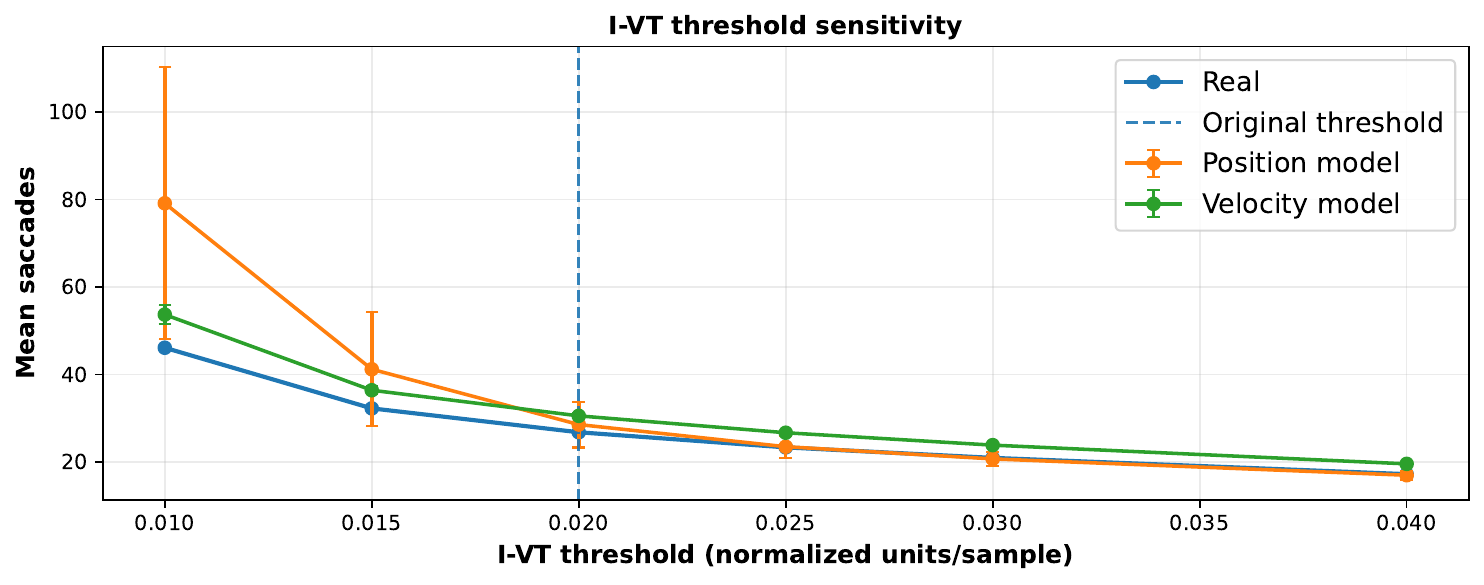}
\caption{I-VT threshold sensitivity. Mean saccade count per eight-second window across 1{,}136 held-out trajectories at six step-speed thresholds. Generated results are averaged over three training seeds; error bars are training-seed SDs. The dashed line marks $0.020$ normalized units/sample. Thresholds span $0.010$--$0.040$, and the counts are operational event labels without a validated degrees/s conversion.}
\label{fig:ivt_threshold_sensitivity}
\end{figure*}

\paragraph{Sampling ablations and event statistics.} Table~\ref{tab:appendix_ddim_ablation} examines the effects of the DDIM sampling budget, sampling stochasticity, and EMA weights for the single reference model. Increasing the number of DDIM steps generally improves distributional agreement, although the improvement from $100$ to $200$ steps is modest relative to the additional sampling cost. We therefore use $100$ steps as the quality--efficiency compromise. Deterministic sampling ($\eta=0$) gives better agreement than increasingly stochastic sampling, and the EMA shadow model reduces the mean-speed KS statistic from $0.277$ to $0.158$ relative to the raw checkpoint. Because these are within-run ablations, they demonstrate sensitivity to the sampling configuration but do not measure variability across independently trained models. Figure~\ref{fig:ivt_threshold_sensitivity} shows the dependence of count on I-VT threshold. Generated results are averaged over three training seeds, with error bars showing training-seed standard deviations.
\begin{table}
\centering
\caption{Nearest-neighbour diagnostics: ranges across three training seeds. Close fractions use the fifth-percentile real-reference distance threshold. Coverage is the median training-to-generated distance divided by the median training-to-validation distance.}
\label{tab:nearest_neighbours}
\footnotesize
\setlength{\tabcolsep}{3pt}
\begin{tabular}{llcc}
\hline
Descriptor & Measure & Position & Velocity \\
\hline
Relative pos. & Close (\%) & 3.52--4.14 & 3.87--4.23 \\
Step vel. & Close (\%) & 3.26--5.28 & 1.06--1.32 \\
Relative pos. & Coverage & 1.100--1.142 & 1.032--1.035 \\
Step vel. & Coverage & 0.978--1.026 & 1.037--1.044 \\
\hline
\end{tabular}
\end{table}

\begin{table}
\centering
\caption{Position-space sampling ablations for the reference run. The DDIM-step and EMA rows report KS/JS for mean speed. The $\eta$ rows report the mean over the eight non-redundant features. Lower is better.}
\label{tab:appendix_ddim_ablation}
\begin{tabular}{lcc}
\toprule
\textbf{Setting} & \textbf{KS} & \textbf{JS} \\
\midrule
DDIM steps = 10            & 0.191 & 0.0565 \\
DDIM steps = 25            & 0.211 & 0.0348 \\
DDIM steps = 50            & 0.180 & 0.0243 \\
DDIM steps = 100 (chosen)  & \textbf{0.158} & \textbf{0.0186} \\
DDIM steps = 200           & 0.151 & 0.0140 \\
\midrule
$\eta=0.0$                 & \textbf{0.135} & \textbf{0.0135} \\
$\eta=0.2$                 & 0.162 & 0.0168 \\
$\eta=0.5$                 & 0.257 & 0.0528 \\
$\eta=1.0$ (stochastic DDIM) & 0.356 & 0.1015 \\
\midrule
Raw model (no EMA)         & 0.277 & -- \\
EMA shadow model (chosen)  & \textbf{0.158} & -- \\
\bottomrule
\end{tabular}
\end{table}
\begin{table}
\centering
\caption{Participant sensitivity under a separate common KDE evaluator. Difference in eight-feature mean JS (velocity minus position) at full $n=1{,}136$, and descriptive 2.5--97.5\% ranges from 200 cluster draws over five validation participants. Generated outputs are fixed. These are not population confidence intervals.}
\label{tab:fun}
\begin{tabular}{lcc}
\toprule
Seed & Difference & Resampling range \\
\midrule
42 & $0.0188$ & $[0.0082,0.0236]$ \\
123 & $0.0257$ & $[0.0136,0.0304]$ \\
2024 & $-0.0018$ & $[-0.0098,0.0043]$ \\
\bottomrule
\end{tabular}
\end{table}
\end{document}